\documentclass[sigconf]{acmart}

\AtBeginDocument{%
\providecommand\BibTeX{{%
\normalfont B\kern-0.5em{\scshape i\kern-0.25em b}\kern-0.8em\TeX}}}

\setcopyright{acmcopyright}
\copyrightyear{2023}
\acmYear{2023}
\acmDOI{XXXXXXX.XXXXXXX}

\acmConference[KnowledgeNLP-KDD'23]{KDD'23 Workshop on Knowledge Augmented Methods for NLP}{August 07, 2023}{Long Beach, CA}

\acmBooktitle{KnowledgeNLP-KDD'23: KDD Workshop on Knowledge Augmented Methods for NLP, August 07, 2023, Long Beach, CA} 
\acmPrice{15.00}
\acmISBN{978-1-4503-XXXX-X/18/06}

\usepackage[utf8]{inputenc} 
\usepackage[T1]{fontenc}    
\usepackage{hyperref}       
\usepackage{url}            
\usepackage{booktabs}       
\usepackage{amsfonts}       
\usepackage{nicefrac}       
\usepackage{microtype}      
\usepackage{xcolor}         
\usepackage{amsthm}
\usepackage{amsmath}
\usepackage[center]{subfigure}
\usepackage{graphicx}
\usepackage{graphbox}
\usepackage{tabularx}
\usepackage{multirow}
\usepackage{threeparttable}
\usepackage{adjustbox}
\usepackage{xspace}
\usepackage{enumitem}
\usepackage{xcolor,colortbl}
\usepackage[]{algorithmic} 
\usepackage{wrapfig}
\usepackage{tablefootnote}
\usepackage{makecell}
\usepackage[font={small}]{caption}
\usepackage{diagbox}
\usepackage{empheq}
\usepackage{todonotes}
\usepackage{tcolorbox}

\usepackage[hang,flushmargin]{footmisc}
\newcommand\workshopnote[1]{\renewcommand\thefootnote{}\footnote{#1}}

\begin{document}

\title{Automatic Question-Answer Generation for Long-Tail Knowledge}
\author{
	Rohan Kumar$^*$,~
	Youngmin Kim$^*$,~
	Sunitha Ravi$^*$,~\\
	Haitian Sun,~
        Christos Faloutsos,~
	Ruslan Salakhutdinov,~
	Minji Yoon$^\dagger$
}
\affiliation{
	\institution{Carnegie Mellon University, Pittsburgh, PA, USA}
        \country{}
}
\email{{rohankum,youngmik,selvansr,haitians,christos,rsalakhu,minjiy}@andrew.cmu.edu}

\renewcommand{\shortauthors}{Kumar, Kim, and Ravi, et al.}

\begin{abstract}
Pretrained Large Language Models (LLMs) have gained significant attention for addressing open-domain Question Answering (QA). 
While they exhibit high accuracy in answering questions related to common knowledge, LLMs encounter difficulties in learning about uncommon long-tail knowledge (tail entities). 
Since manually constructing QA datasets demands substantial human resources, the types of existing QA datasets are limited, leaving us with a scarcity of datasets to study the performance of LLMs on tail entities. 
In this paper, we propose an automatic approach to generate specialized QA datasets for tail entities and present the associated research challenges. 
We conduct extensive experiments by employing pretrained LLMs on our newly generated long-tail QA datasets, comparing their performance with and without external resources including Wikipedia and Wikidata knowledge graphs.
\end{abstract}


\maketitle
\workshopnote{
$^*$These authors contributed equally. \\ 
$^\dagger$ Corresponding author: <minjiy@andrew.cmu.edu> \\
Accepted to Second Workshop on Knowledge Augmented Methods for Natural Language Processing, in conjunction with KDD 2023.}

\section{Introduction}
\label{sec:introduction}

Open-domain Question Answering (QA)~\cite{joshi2017triviaqa, kwiatkowski2019natural}, which involves answering questions regarding common knowledge, has long been a challenging task in the fields of natural language understanding, information retrieval, and related domains~\cite{voorhees1999trec, moldovan2000structure}. 
Large language models (LLMs) trained on internet text effectively capture a wide range of world knowledge, encompassing both widely known facts and domain-specific information. 
These models have achieved remarkable success in QA tasks, eliminating the need for external documents during inference by implicitly storing knowledge in their parameters~\cite{petroni2019language, roberts2020much, kandpal2022large}.

However, the impressive achievements of LLMs in QA tasks are primarily observed with regard to common concepts that frequently appear on the internet (referred to as "head entities"), which are thus more likely to be learned effectively by LLMs during pre-training time. 
Conversely, when it comes to dealing with long-tail knowledge, which encompasses rarely occurring entities (referred to as "tail entities"), LLMs struggle to provide accurate answers and often exhibit hallucination issues ~\cite{ji2023survey}.
Due to the predominant focus of most QA datasets on head entities~\cite{kwiatkowski2019natural, joshi2017triviaqa, dunn2017searchqa}, research investigating the performance of LLMs on long-tail knowledge has been limited.
Recently, \citet{kandpal2022large} conducted a study to bridge this gap by constructing dedicated QA datasets for tail entities. 
Their approach involved leveraging the entity frequency in Wikipedia to define tail entities and quantitatively demonstrating the limitations of LLMs in handling such entities.

Wikipedia documents~\cite{lee2019latent} and Wikidata knowledge graphs~\cite{vrandevcic2014wikidata} are the primary external resources from which QA models acquire knowledge. 
Consequently, the distribution of tail entities is largely determined by the knowledge distributions within Wikipedia and Wikidata. 
In this study, we propose a novel approach to defining tail entities based on their degree information in Wikidata, as opposed to~\cite{kandpal2022large} relying on Wikipedia.
By doing so, we generate QA datasets with distinct distributions from previous works~\cite{kandpal2022large}, thus fostering diversity within tail-knowledge QA datasets.
Within the context of Wikidata, the degrees of entities reflect their level of engagement with general knowledge. 
Hence, we leverage this degree information to define tail entities.

The construction of QA datasets typically requires significant human resources, hindering the creation of diverse datasets from various domains that are essential for testing the robustness of current QA models. 
In this study, our main emphasis lies on the \textit{automatic generation} of long-tail QA datasets. 
However, we encounter several challenges in this process, such as filtering noisy questions, question granularity, difficulty control, and prompt engineering. 
These challenges necessitate further research to identify fundamental solutions. 
We present these challenges through insightful case studies, aiming to stimulate additional research in this area and foster the development of QA models.

Lastly, we assess the performance of pretrained LLMs, specifically GPT3, on our tail entity datasets. 
Our findings reveal distinct patterns compared to prior work~\cite{kandpal2022large}, which defines tail entities based on Wikipedia rather than Wikidata. 
This underscores the importance of utilizing diverse QA sets to accurately gauge the robustness of QA models. 
Moreover, we investigate strategies to enhance the performance of pretrained LLMs by incorporating external resources, such as external documents or knowledge graphs, during inference time on our automatically-generated long-tail QA datasets. 
We link these experimental results and the challenges encountered during the automatic QA dataset generation process. 
In summary, our contributions encompass:
\begin{itemize}[leftmargin=10pt,topsep=1pt,itemsep=-1ex,partopsep=1ex,parsep=1ex]
    \item {Introduction of novel tail knowledge QA datasets derived from the Wikidata knowledge graph.}
    \item {Outline of the automatic construction of QA datasets and associated open research challenges.}
    \item {Evaluation of GPT3's performance with/without external resources on our new datasets.} 
\end{itemize}

\noindent Our code is publicly available$^{1}$.~\footnote{$^{1}$ \url{ https://github.com/isunitha98selvan/odqa-tail}}

\section{Related Work}
\label{sec:related_work}

\textbf{Fact Learning by LLMs}: 
LLMs \cite{brown2020language, zhang2022opt, touvron2023llama} have shown state-of-the-art performance across various NLP tasks. 
LLMs have been shown to memorize facts successfully by learning high-frequency patterns in the training data~\cite{tanzer2021memorisation, kassner2020pretrained, kandpal2022large}. 
\citet{kandpal2022large} show that an LLM's ability to answer a question is affected by how many times it has seen relevant documents related to the question in its pre-training data. They show that LLMs struggle to reason accurately over rarer entities in the pre-training data (ROOTS \cite{laurenccon2022bigscience}, C4 \cite{raffel2020exploring}, Wikipedia \cite{lee2019latent}, OpenWebText \cite{gokaslan2019openwebtext}). In this work, instead of using the pre-training corpus, we define tail entities using Wikidata knowledge graphs and construct a long-tail knowledge dataset that can be used to study the open-domain QA performance of LLMs. 


\noindent \textbf{Open-domain Question Answering}: 
ODQA is widely used to measure the fact-learning performance of LLMs.
However, most of them are composed of common knowledge (i.e., head-entity questions), which prevents deep investigations into LLM’s ability to learn facts about uncommon concepts.
For instance, TriviaQA \cite{joshi2017triviaqa} is generated from trivia websites, where questions are generally about popular entities or facts. Similarly, NaturalQA \cite{kwiatkowski2019natural} is constructed manually using queries issued to the Google search engine. One reason that it is hard to find tail-entity datasets is, most of the time, QA datasets are hand-crafted, requiring a large number of human resources; thus, the types of QA datasets are limited to certain types (e.g., head entities).
Here, we focus on how to generate tail-entity datasets while minimizing human resources and analyze why the automatic QA dataset construction is nontrivial.

\begin{figure}
    \centering
    \includegraphics[width=9cm,height=6cm,keepaspectratio]{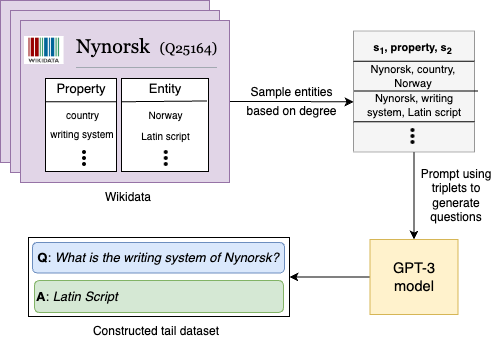}
    \caption{
    Overview of the automatic QA data construction process for long-tail knowledge: 
    We first sample tail entities that have low degrees and extract the connected triplets from Wikidata knowledge graph;
    Then we prompt GPT3 with the triplets to generate natural language questions.
    }
    \label{fig:overview}
\end{figure}

\section{Automatic Generation of QA datasets for Long-Tail Knowledge}
\label{sec:qa_generation}

In this section, we first describe how to generate QA datasets from Wikidata knowledge graphs automatically, then list associated challenges in the process. 

\subsection{Overview}
\label{sec:qa_generation:overview}

In knowledge graphs, a triplet [\emph{s1, property, s2}] consists of a subject entity node \emph{s1}, an edge \emph{property}, and an object entity node \emph{s2}. 
A triplet represents a piece of information about \emph{s1} that can be used to generate a question/answer pair about \emph{s1}. 
For instance, a triplet [\emph{The Hospital, location, New York City}] represents the information that \emph{The Hospital} is \emph{located} in \emph{New York City}.

We define tail entities based on each entity's node degree (i.e., the number of triplets that have the target entity as a subject node \emph{s1}) in the knowledge graph. 
We first sample tail entities based on their degree information and extract all triplets that have the tail entities as the subject entity from Wikidata (proper degree bounds of tail entities will be discussed in the following section).
Then we generate factoid questions by prompting LLMs with triplets. 
Specifically, we use the GPT3 model with the following prompt:
\begin{tcolorbox}
\begin{verbatim}
Generate questions:
obama | born | hawaii => where was obama born?
sky | color | blue => what color is the sky?
X | Y | Z => 
\end{verbatim}
\end{tcolorbox}
Here X, Y, and Z correspond to \emph{s1, property, s2} of triplets extracted from the Wikidata.
Figure~\ref{fig:overview} outlines how to automatically generate QA datasets for tail entities from knowledge graphs.

\subsection{Challenges}
\label{sec:qa_generation:challenges}

In this section, we describe open research challenges we faced while constructing QA datasets automatically from knowledge graphs.

\begin{figure}[h]
    \centering
    \includegraphics[width=0.8\linewidth]{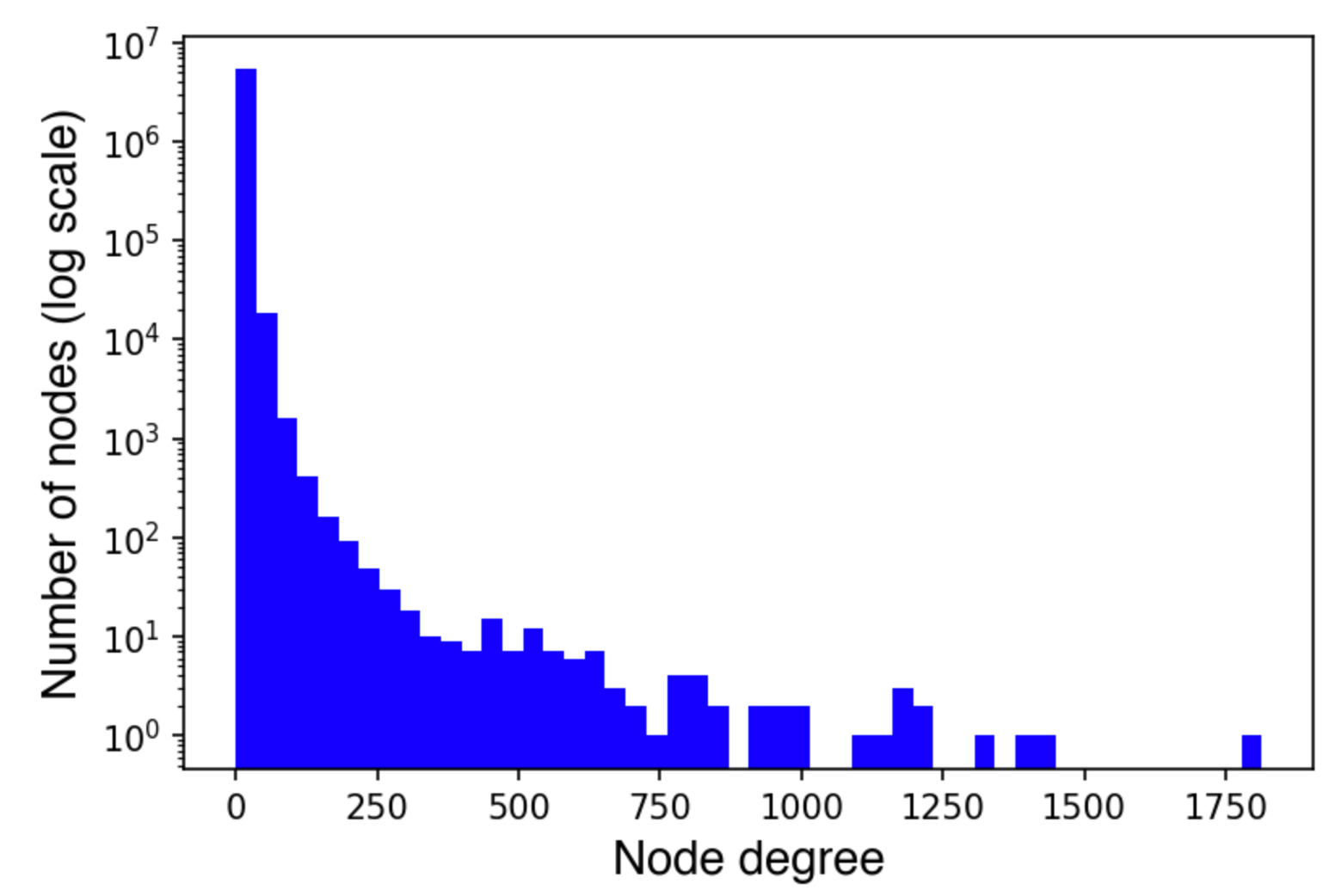}
    \caption{Node degree distribution of all entities in Wikidata.}
    \label{fig:node_degree_dist}
\end{figure}

\subsubsection{Degree bounds for tail entities}
\label{sec:qa_generation:challenges:definition}

There are no strictly-formulated definitions for tail entities that are widely accepted. 
Degree bounds that instantly bring in differences in model performance are also hard to be decided in advance.
As a result, degree bounds for tail entities should be selected arbitrarily. 
In our experiments, we classify entities with node degrees between $15$ and $100$ as \textit{coarse}-tail entities and entities with node degrees below $3$ as \textit{fine}-tail entities and compare the LLM performance on them.
Figure~\ref{fig:node_degree_dist} shows the degree distribution of entities in Wikidata.

\subsubsection{Filtering out noisy triplets}
\label{sec:qa_generation:challenges:ambiguous_triplets}

\paragraph{\textbf{Ambiguous entities:}} 
Multiple entities can have the same surface forms. 
For instance, \textit{Jesus} can refer to either \textit{1999 Biblical telefilm directed by Roger Young}, \textit{1979 film by Peter Sykes}, or \textit{central figure of Christianity} in Wikidata.
While these entities can be distinguished by their unique IDs in knowledge graphs, questions generated from such entities are ambiguous to answer (e.g., \textit{Who was cast in Jesus?}).
This introduces complications in evaluating the correctness of a model's answers. 
Similarly, answer entities can have several correct surface forms. 
\textit{World War II} can be written as \textit{WW2} , \textit{WWII}.  
In our experiment, we use correct answers along with their aliases from Wikidata for evaluation.
However, aliases provided by Wikidata do not cover all possible surface forms, leading to high false negatives.

\paragraph{\textbf{Ambiguous properties:}}
In Wikidata, a large number of properties cannot be used to generate sensible questions.
For instance, \textit{subclass of}, \textit{instance of}, or \textit{part of} would generate questions that are too vague to answer even for humans.
Another example is \textit{family name}, which will generate questions that already contain the answer in them (e.g., \textit{What is the family name of Barack Obama?}). 
Wikidata also has properties that merely link entities to images or URLs, such as \textit{logo image}, \textit{official website}, and \textit{official blog URL}. 
Questions generated with these properties are not helpful in evaluating QA model performance, so they need to be filtered out.

As there is no straightforward metric to quantify the appropriateness of each property for question generation, property filtering is difficult to be automated. 
Property filtering requires human judgment, which can be problematic because it can be subjective as well as difficult to scale.
In our experiment, we filter out properties by manually going through all the properties that are initially extracted.

\subsubsection{Difficulty control}
\label{sec:qa_generation:challenges:difficulty_control}

\begin{figure}[h]
    \centering
    \includegraphics[width=0.7\linewidth]{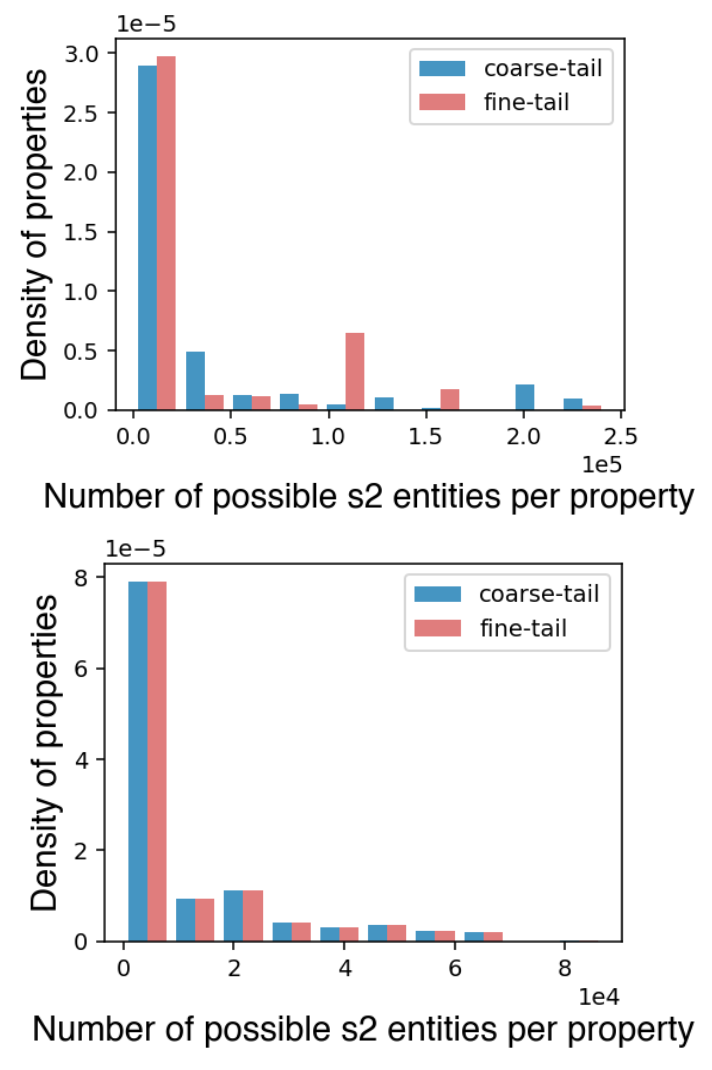}
    \caption{Density of properties per the number of possible \textit{s2} object entities before (Top) and after (Bottom) the difficulty controlling.}
    \label{fig:dist_change}
    \vspace{-3mm}
\end{figure}

Questions generated from different properties can have different levels of difficulty.
For example, the property \textit{driving side} only has two possible choices, \textit{right} and \textit{left}, for the object (answer) entity.
In contrast, the property \textit{child} has approximately $800k$ possible choices for the object entity in Wikidata.
The difficulty of selecting the correct answer for these two properties can therefore be very different. 

Our goal is to generate long-tail QA sets based on the degree of subject (question) entities in knowledge graphs.
In other words, we want the difficulty of questions solely affected by the degree of question entities, not by properties.
In Figure~\ref{fig:dist_change}, we match \textit{coarse}-tail dataset and \textit{fine}-tail dataset to contain the same number of triplets for each property, normalizing the difficulty of QA sets in terms of properties.

\subsubsection{LLM prompt for question generation}
\label{sec:qa_generation:challenges:llm_prompt}

While the answer entity of a triplet is not part of the generated question, we find that the quality of generated questions improves when the complete triplet is provided in the prompt, instead of the first two elements (i.e., subject entity and property). 
For instance, given a triplet [\textit{david peel yates, conflict, world war ii}], we get \textit{"What conflict was David Peel Yates involved in?"} from GPT3 when using just the subject entity and property in prompt. 
On the contrary, when we use all subject, property, and object entities, the generated question becomes \textit{"What conflict did David Peel Yates serve in?"}.
By including the answer entity \textit{world war ii} in the prompt, GPT3 understands \textit{conflict} is about a war, not about people, and generates a question with a more proper verb \textit{serve in}.
This shows that prompt engineering for generative LLMs is crucial for the quality of generated QA datasets.

\subsubsection{Granularity of questions}
\label{sec:qa_generation:challenges:answer_granularity}

Given a question, there could be several correct answers with different granularity.
Unless the question specifies the granularity of the answer (e.g., \textit{which country} or \textit{which city}), QA datasets and models could easily pick different granularity of answers.
For instance, when asked \textit{Where was Lovelyz formed?}, a model could answer \textit{South Korea} while the QA dataset has \textit{Seoul} (the capital of South Korea) as the correct answer and marks the predicted answer wrong. 
To specify the granularity of the answer in the question, generative LLMs should already know about the question/answer entity, which becomes problematic if these entities are long-tail knowledge.
The other solution is preparing all possible granularity as the correct answers, which is also practically infeasible.

\begin{table*}[]
    \caption{
        Error analysis on \textit{Fine}-tail QA set.
        }
    \label{tab:llm-error}
    \centering
    \begin{tabular}{l|l|r}\toprule\hline 
    \textbf{Reason} & \textbf{Explanation} & \textbf{Ratio} \\ \midrule\hline
    Incorrect & Wrong answers & 45\% \\
    Granularity & Answers are too specific or too generic & 19\% \\
    Incorrect question & Unrelated to input triplets & 12\% \\
    Exact match & Answers are correct but don't exactly match (e.g., no punctuation, synonyms) & 9\% \\
    Multiple answers & Both answers and predictions are correct, but questions have multiple answers & 3\% \\
    Others & e.g., input triplets are not useful/ambiguous & 12\% \\ \hline\bottomrule
    \end{tabular}
\end{table*}

\section{Evaluation with LLMs and external resources}
\label{sec:evaluation}

\subsection{Experiment Setup}
\label{appendix:exp_setting}

\textbf{GPT3:} We use the \textit{text-davinci-003} version of GPT3 for all our experiments (e.g., question generation, answering questions). 
The model is accessible via the OpenAI API$^{2}$\footnote{$^{2}$ https://platform.openai.com/docs/api-reference/introduction}. Specifically, we make use of the Completions API to prompt the model.\\
\textbf{Wikipedia:} 
We use Wikipedia articles to retrieve relevant paragraphs on-the-fly to augment the GPT3 prompt with additional context. 
To find the relevant paragraphs using Dense Passage Retriever (DPR)~\cite{karpukhin2020dense}, we use the same Wikipedia dump as in the original paper~\cite{karpukhin2020dense}.\\
\textbf{Wikidata:} Wikidata knowledge graph consists of $103,305,143$ entities and $11,007$ properties. We access Wikidata using the Sling tool \cite{ringgaard2017sling} in a triplet format (\textit{subject}, \textit{property}, \textit{object}).\\
\textbf{Tail-entity datasets:} 
We sample triplets from Wikidata to create \textit{Coarse}-tail and \textit{Fine}-tail datasets.
Each dataset has $27,691$ triplets and $422$ unique properties after the difficulty control (details in Section~\ref{sec:qa_generation:challenges:difficulty_control}). 
One question$\&$answer pair consists of a GPT3-generated question, an answer (i.e., object entity in the original triplet), and associated aliases for the answer.

\subsection{LLM prompting for open-domain QA}
\label{sec:evaluation:llm}

We study the performance of GPT3 on our tail-entity datasets.
We prompt the GPT3 model with few QA pairs as follows:
\begin{tcolorbox}
\begin{verbatim}
Answer the given question:
where was obama born? => hawaii
what color is the sky? => blue
where was lovelyz formed? => 
\end{verbatim}
\end{tcolorbox}
Table \ref{tab:llm} shows GPT3 performance on existing QA datasets (TriviaQA~\cite{joshi2017triviaqa}, WebQA~\cite{berant2013semantic}, and NaturalQA~\cite{kwiatkowski2019natural}) and our newly-generated \textit{Coarse-} and \textit{Fine-}tail QA datasets.
GPT3 shows consistently lower performance on our tail datasets than the existing QA datasets, while performing better on \textit{Coarse-}tail set than \textit{Fine-}tail set.
This results coincide with \cite{kandpal2022large}, showing again that LLMs struggle to learn long-tail knowledge.

\begin{table}[h]
    \caption{
        GPT3 few-shot performance on open-domain QA datasets.
        $^*$Results for TriviaQA, WebQA, and NaturalQA are from \cite{brown2020language}.
        }
    \centering
    \begin{tabular}{l|c}\toprule\hline
        Dataset & Accuracy \\ \midrule\hline
        TriviaQA$^*$ &  71.2\% \\
        WebQA$^*$ &  41.5\% \\
        NaturalQA$^*$ & 29.9\% \\ \hline
        \textit{Coarse}-tail QA & 26.5\% \\
        \textit{Fine}-tail QA & 22.1\% \\ \hline\bottomrule
    \end{tabular}
    \label{tab:llm}
\end{table}

We perform manual error analysis on our tail QA dataset.
We randomly sample $100$ questions that got wrong from \textit{Fine}-tail QA set and categorize their error cases into $6$ cases.
As shown in Table~\ref{tab:llm-error}, $45\%$ of errors are from GPT3's completely wrong predictions. 
$19\%$ of errors are due to different granularity of answers, and $12\%$ of errors are due to questions that are incorrectly generated by LLMs.
As we describe in Section~\ref{sec:qa_generation:challenges}, this result shows the limitations of auto-generated QA datasets and underscores the imperative for further research in this domain.

\subsection{LLM prompting with Dense Passage Retrieval}
\label{sec:evaluation:llm_dpr}

One common way to augment LLMs for long-tail knowledge is retrieving relevant passages from external documents and referring them during  inference time~\cite{kandpal2022large}. 
In this section, we check whether GPT3 can see the same improvement in our datasets.
We use Dense Passage Retriever (DPR)~\cite{karpukhin2020dense} that has trained on Natural Questions~\cite{kwiatkowski2019natural} to retrieve the top $100$ relevant passages from Wikipedia.

We first evaluate how successfully DPR retrieves a passage that contains the correct answer.
In Table~\ref{table:dpr}, we observe that DPR performs consistently worse on our tail datasets than the existing datasets. 
This shows that DPR, which has pretrained on Wikipedia with head-entity QA datasets, also struggles to retrieve long-tail knowledge.

\begin{table}[h]
    \caption{
        Performance of DPR on retrieving relevant documents from Wikipedia.
        We check Top-$K$ retrieved passages contain the correct answer.
        $^*$Results for TriviaQA, WebQA, and NaturalQA are from \cite{karpukhin2020dense}.
        }
    \label{tab:dpr}
    \centering
    \begin{tabular}{l|cc}\toprule\hline
    Dataset & Top-$20$ & Top-$100$ \\ \midrule\hline
    TriviaQA$^*$ & 79.4\% &  85.0\% \\
    WebQA$^*$ & 73.2\% &   81.4\% \\
    NaturalQA$^*$ & 78.4\% & 85.4\% \\ \hline
    \textit{Coarse}-tail QA & 50.5\% & 63.3\% \\
    \textit{Fine}-tail QA & 54.5\% & 66.3\% \\ \hline\bottomrule
    \end{tabular}
    \label{table:dpr}
\end{table}

Next, we use the top-ranked passages retrieved by DPR to augment GPT3.
We pass the top-$1$ retrieved passages to GPT3 as additional context along with the question as follows:

\begin{tcolorbox}
\begin{verbatim}
Question: Where is Nelson's Pillar located?
Document: Nelson's Pillar was a large granite 
column capped by a statue of Horatio Nelson, 
built in the centre of what was then Sackville 
Street in Dublin, Ireland.
Answer: Dublin, Ireland 
\end{verbatim}
\end{tcolorbox}

In Table ~\ref{table:dpr-llm}, we observe a decrease in GPT3 accuracy compared to its original accuracy when prompted with DPR retrieved passages.
The accuracy of $26.5\%$ for \textit{Coarse}-tail and $22.1\%$ for \textit{Fine}-tail QA sets plummet to $14.3\%$ and $18.2\%$, respectively.
This decline in accuracy can be attributed to the fact that DPR's retrieval often leads to irrelevant passages on long-tail knowledge, as shown in Table~\ref{table:dpr}. 
Consequently, the presence of these additional contexts confuses GPT3 and adversely affects its performance. 
These findings highlight the crucial relationship between the performance of LLMs and the retrieval models, indicating that the performance of LLMs is inherently limited by the effectiveness of the retrieval models. 
Therefore, it is essential for retrieval models to also consider and address the challenges associated with long-tail knowledge.

\begin{table}[h]
    \caption{
        Performance of GPT3 prompted with the Top-$1$ DPR retrieved passage.
        DPR's low accuracy leads to irrelevant passages being retrieved.
        Then additional contexts confuse GPT3, leading to a decrease in accuracy compared to its original performance.
    }
    \label{table:dpr-llm}
    \centering
    \begin{tabular}{l|c|c} \toprule\hline
      & \textit{Coarse}-tail QA & \textit{Fine}-tail QA \\ \hline
     Original & 26.5\%  & 22.1\% \\
     w/ DPR & 14.3\%  & 18.2\% \\ \hline\bottomrule
     \end{tabular} 
\end{table}

\subsection{LLM prompting with DPR and knowledge graphs}
\label{sec:evaluation:llm_dpr_kg}

Knowledge graphs (KG) have been widely used to augment LLMs~\cite{zhang2022greaselm, sun2021jointlk}.
In this section, we examine how external knowledge graphs can cooperate with another external resource, Wikipedia to improve LLM performance for tail entities. 
We use Wikidata as our external knowledge graph after removing all triplets used for the QA generation.
To avoid additional finetuning, we implement a zero-shot LLM+DPR+KG baseline: 
we first sample triplets relevant to the question from the knowledge graph then use the sampled triplets to rerank the DPR-retrieved passages; then we pass the top-$1$ retrieved passages to GPT3 as additional context along with the question.
To sample relevant triplets from knowledge graphs, we first find a path from the subject entity to the object entity and concatenate the surface forms of all entities on the path. 
We then compute its textual similarity with the passages retrieved by DPR using SBERT~\cite{reimers2019sentence}. 
We use this similarity score to re-rank the DPR results and observe the changes in the Top-$K$ retrieval accuracy.

\begin{table}[h]
    \caption{
        Top-$K$ DPR retrieval accuracy and GPT3 performance on \textit{Fine}-tail QA before/after the retrieved passage re-ranking using knowledge graphs.
        The third and final columns (\textit{re-rank}) show how Top-$K$ DPR retrieval accuracy and GPT3 performance have changed after the re-ranking.
        }
    \label{tab:llm_dpr_kb}
    \centering
\begin{tabular}{l|c|c|c|c}\toprule\hline
    & DPR & re-rank & \textit{Fine}-tail QA& re-rank \\ \midrule\hline
 Top-$1$ & 23.04\% & 29.10\% & \multirow{4}{*}{22.10\%} & \multirow{4}{*}{30.95\%} \\
 Top-$20$ & 59.56\% & 65.13\% & &  \\
 Top-$50$ & 69.99\% & 72.44\% &  &  \\
 Top-$100$ & 75.65\% & 75.65\% &  &  \\ \hline\bottomrule
\end{tabular}
\end{table}

As shown in Table~\ref{tab:llm_dpr_kb}, DPR retrieval accuracy is improved by up to $6\%$ with the help of knowledge graphs.
The improvement in DPR retrieval accuracy leads to the improvement of GPT3's QA performance.
Table~\ref{tab:llm_dpr_kb} shows GPT3 also has improved from $22.1\%$ (no external resources) to $30.95\%$ (with DPR and knowledge graphs) on our \textit{Fine}-tail QA datasets.
This highlights that joint learning of two external resources could be the key to solving the long-tail knowledge problems.

\section{Conclusion}
\label{sec:conclusion}

Our work highlights the limitations of pre-trained LLMs in handling long-tail knowledge in open-domain Question Answering.
To investigate this limitation, we first propose to generate QA datasets specialized for tail entities automatically using degree information from the Wikidata knowledge graph. 
Our automatic QA generation approach aims to overcome the resource-intensive nature of manual dataset construction, allowing for the creation of diverse long-tail QA datasets.
In the process of automatic QA dataset generation, we identify and discuss several open research challenges, such as degree bounds, question granularity, difficulty control, and prompt engineering, which require further investigation for fundamental solutions. 
We evaluate the performance of GPT3 on our generated long-tail QA datasets. 
Additionally, we explore the utilization of external resources, such as external documents or knowledge graphs, to improve the performance of LLMs on long-tail knowledge. 
We hope this work paves the way for further research in the automatic QA dataset generation and the long-tail knowledge problem in open-domain QA tasks.

\bibliographystyle{ACM-Reference-Format}
\bibliography{custom}

\end{document}